\documentclass[a4paper,fleqn]{cas-dc}

\usepackage[authoryear,longnamesfirst]{natbib}
\usepackage{caption}
\usepackage{subcaption}

\def\tsc#1{\csdef{#1}{\textsc{\lowercase{#1}}\xspace}}
\tsc{WGM}
\tsc{QE}
\tsc{EP}
\tsc{PMS}
\tsc{BEC}
\tsc{DE}

\begin{document}
\let\WriteBookmarks\relax
\def\floatpagepagefraction{1}
\def\textpagefraction{.001}

\shorttitle{FRF4POD}

\shortauthors{Park et~al.}

\title [mode = title]{Federated Random Forest for Partially Overlapping Clinical Data}                      



%
\author[1]{Youngjun Park}[type=editor,
                        auid=000,bioid=1,
                        ]






\affiliation[1]{organization={Department of Medical Informatics, University Medical Center Göttingen},
    city={Göttingen},
    country={Germany}}

\author[1]{Cord Eric Schmidt}

\author[2]{Benedikt Marcel Batton}[%
   ]


\affiliation[2]{organization={Department of Mathematics and Computer Science, Philipps-Universität Marburg},
    city={Marburg},
    country={Germany}}

\author[1]{Anne-Christin Hauschild}[type=editor,
                        auid=000,bioid=1,
                        orcid=0000-0002-7499-4373,
                        ]
\cormark[1]
\ead{anne-christin.hauschild@med.uni-goettingen.de}

\cortext[cor1]{Corresponding author}



\begin{abstract}
In the healthcare sector, a consciousness surrounding data privacy and corresponding data protection regulations, as well as heterogeneous and non-harmonized data, pose huge challenges to large-scale data analysis.
Moreover, clinical data often involves partially overlapping features, as some observations may be missing due to various reasons, such as differences in procedures, diagnostic tests, or other recorded patient history information across hospitals or institutes. To address the challenges posed by partially overlapping features and incomplete data in clinical datasets, a comprehensive approach is required.
Particularly in the domain of medical data, promising outcomes are achieved by federated random forests whenever features align. However, for most standard algorithms, like random forest, it is essential that all data sets have identical parameters.
Therefore, in this work the concept of federated random forest is adapted to a setting with partially overlapping features. 
Moreover, our research assesses the effectiveness of the newly developed federated random forest models for partially overlapping clinical data. 
For aggregating the federated, globally optimized model, only features available locally at each site can be used.
We tackled two issues in federation: (i) the quantity of involved parties, (ii) the varying overlap of features.
This evaluation was conducted across three clinical datasets.
The federated random forest model even in cases where only a subset of features overlaps consistently demonstrates superior performance compared to its local counterpart. This holds true across various scenarios, including datasets with imbalanced classes. 
Consequently, federated random forests for partially overlapped data offer a promising solution to transcend barriers in collaborative research and corporate cooperation.
\end{abstract}




\maketitle

\section{Introduction}
Machine learning models require data of sufficient extent and quality in order to perform adequately. In recent years, privacy awareness has risen, and in several countries, data privacy legislation has been adopted, restricting the exchange of data \cite{yang2019federated}.
Most prominently, the General Data Protection Regulation (GDPR) regulates data privacy in Europe but also applies to foreign companies or persons doing business in Europe. The GDPR prohibits the sharing of data for causes not disclaimed when data is raised \cite{regulation2016regulation}.
This has fuelled the current surge of new developments in the field of federated learning. Federated learning allows to train models without aggregating data in one place by only exchanging models or their parameters. 
The goal is to collaboratively train a federated model which shows similar performance to a model trained on the aggregated data.
The fundamental challenge in federated learning is forming a centrally combined learning model without having access to the underlying data.
Tian \textit{et al.} summarized challenges into four categories: (1) expensive communication, (2) heterogeneous system, (3) data heterogeneity, and (4) privacy \cite{li2020federated}.
In this work, our main focus will be on exploring data heterogeneity.

Real-world datasets exhibit heterogeneity, characterized by skewed data distribution, imbalanced data, and partially overlapping features. 
Especially, clinical data often contains many missing values due to a variety of reasons, including the different procedures in hospitals, the data collection process, the characteristics of the participants, and the design of the study itself \cite{le2020challenges, tong2022distributed}.
These challenges also need to be addressed in the context of federated learning.
To address the challenge of missing values and handle partially overlapping features, a comprehensive approach is required.
One strategy involves addressing missing values with imputation \cite{danks2008integrating, austin2021missing, heymans2022handling}. 
In a federated learning setting, transfer learning or a generative model-based approach is introduced for missing data imputation \cite{yao2022fedtmi, zhou2021federated}.
Another option for handling missing data, besides imputation, is to use models that can inherently handle missing values. 
These models are designed to work with incomplete data and can make predictions or classifications without requiring the missing values to be filled in \cite{emmanuel2021survey}. 

The random forest is an ensemble method that builds multiple decision trees on different subsets of the data and averages their predictions. This method can handle missing values by allowing each tree in the forest to learn from a different subset of the data, potentially capturing different patterns or relationships that include instances with missing values \cite{emmanuel2021survey}.
There are different strategies for building federated random forest model.
Each node independently trains decorrelated decision trees using its local dataset. These locally trained trees are then combined centrally to form the final FRF model \cite{hauschild2022federated, gencturk2022bofrf}. 
Another strategy involves training a global model that aggregates histograms from all participating nodes. Each node computes a histogram over its local dataset, representing the distribution of feature values within that dataset. These histograms are then sent to a central location, where they are aggregated to form a comprehensive view of the entire dataset \cite{kalloori2022cross}.

This work we extended the federated random forest (FRF) approach described in \cite{hauschild2022federated, gencturk2022bofrf}
which have demonstrated superior performance compared to local models on average and has shown comparable performance to data-centralized models trained on the entire dataset. 
As the number of models increases and the dataset size decreases, local models experience a significant decrease in performance, whereas the performance of FRF remains relatively stable. When integrating datasets of varying sizes, the FRF significantly outperforms average local models. Through analyzing different class imbalances, we further confirm the robustness of FRF, which consistently outperforms local models.
This findings suggest that FRF effectively transcends the limitations of clinical research, facilitating collaborations across institutions without compromising privacy or legal compliance. Clinicians stand to benefit from access to an extensive repository of unbiased data sourced from diverse geographic regions, demographics, and other variables. 

This work focuses on the evaluation of the newly developed and advanced FRF on distributed clinical datasets that comprise partially overlapping features.
The primary consideration is to prove the benefit of federated random forest models in a horizontal federated learning setup. 
We compared the globally optimized models to the local models' performances in two scenarios: (1) varying number of sites and (2) varying amount of overlapping features.
The expectations for the federated random forest model are summarized in the following two proposed hypotheses.
Hypothesis 1: The globally optimized federated models outperform local random forest models when applied to clinical data with partially overlapping features.
Hypothesis 2: With less overlap, there is less improvement in model performance between local and globally optimized models. With smaller overlap in features, only partial information is shareable through the model itself.

\section{Material and Methods}

\subsection{Federated random forest for partially overlapping data}

\begin{figure}
    \centering 
    \includegraphics[clip, trim=2.2cm 2cm 1.8cm 2cm, width=0.50\textwidth]{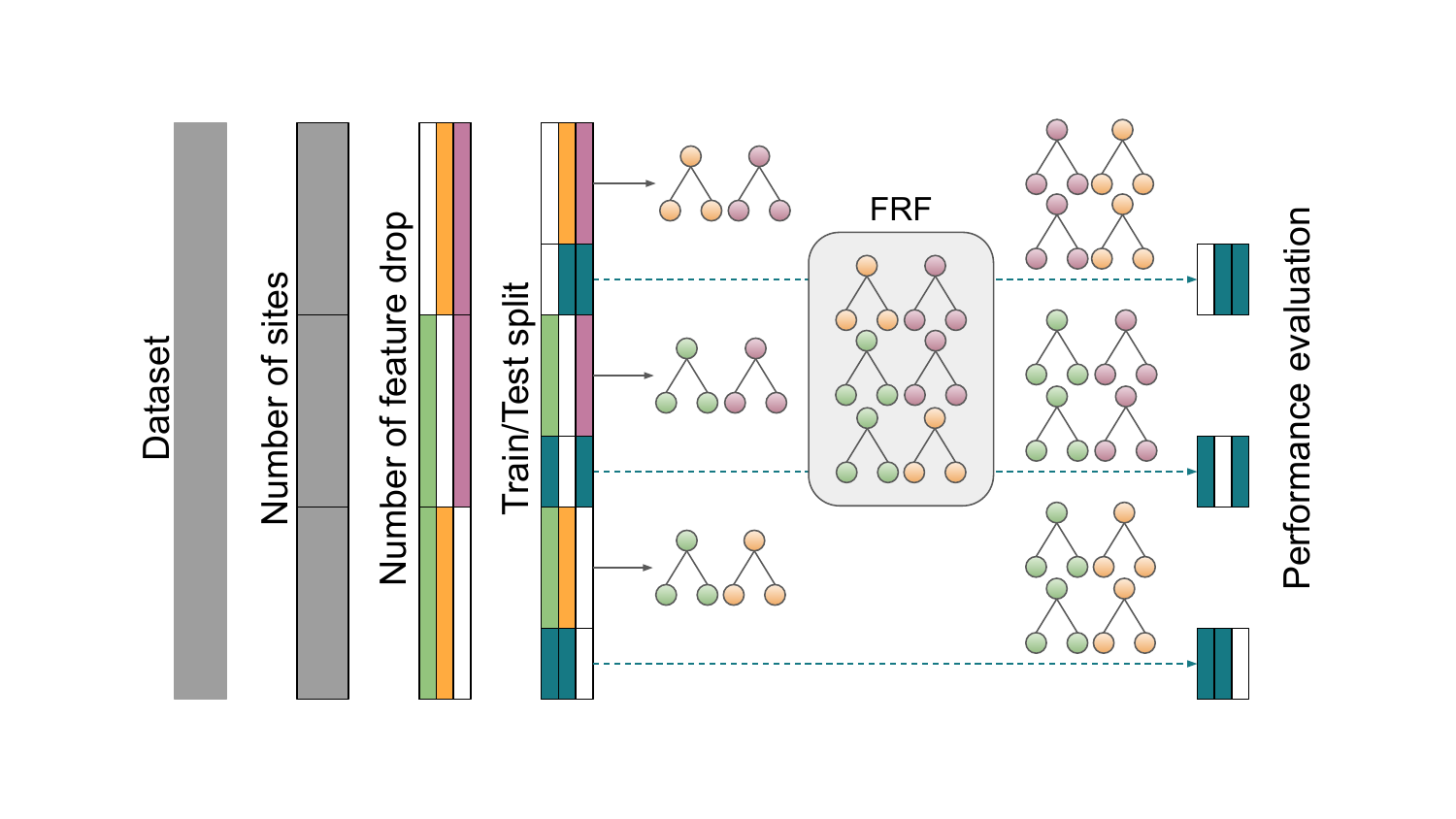}
    \caption{The overview of the suggested federated random forest model. 
    Data is split into multiple sites and random features are dropped to simulate partially overlapping conditions.
    The federated random forest model for partially non-overlapping data aggregate all trees from local random forest models.
    The sites can request and extract a globally aggregated random forest model. Then each optimized local model is evaluated with the respective local test data.
    }
    \label{fig:overview}
\end{figure}

The federated random forest is extended for partially overlapping data to handle issues in real-world data where samples are split across different sites, and the features only partially overlap across different hospitals or institute.
In the first step, each site trains a local random forest (local RF) with the locally available data.
Secondly, to improve the performance of random forest models built locally at each site, the local RFs are aggregated at a central server to form a global random forest (global federated RF) containing all local trees. 
Next, each local site can request the decision tree models from the global federated RF, which only uses locally available features as a split criterion. 
Finally, the globally optimized local random forest (go-local RF) can be applied and evaluated locally. For both ends, locally and globally, dictionaries are used to align the trees to the local data format and to determine which trees can be used for the optimized model based on the locally available features, as these differ between sites. 
Other sites can subsequently benefit from the additional information included in the committed model when requesting their updated, optimized local random forest. 

For creating the optimized model at the different sites, additive and constant aggregation can be used. Using the additive aggregation method, all trees that exploit the features available at the local site are added to the updated model. Therefore, in addition to the locally available trees, trees from other sites which can be used locally are also added. However, for the constant aggregation method, the number of trees should remain the same. To incorporate the information obtained from the other sites into the locally optimized model, a sample holding the same number of trees as the initial local model is drawn from the list of transferable trees. All trees hereby have the same probability of being included in the sample.


\subsection{Data description}

\begin{table}[h]
    \centering
    \caption{Data description used in analysis. The number of samples, features, and class are shown and imbalance ratio (IR) is also shown.}
    \begin{tabular}{c|c c c c}
        \toprule
         Data Set & \#Sample & \#Feature & \#Class & IR \\
         \midrule
         ILPD & 579 & 10 & 2 & 0.4  \\
         HCC & 685 & 7 & 2 & 0.69 \\
         BCD & 569 & 30 & 2 & 0.59  \\
         \bottomrule
    \end{tabular}
    \label{tab:datadescription}
\end{table}

This section presents an overview of the data sets used to implement and evaluate the proposed federated random forest (FRF).
Table \ref{tab:datadescription} outlines the data sets and gives an overview of the number of samples, features, and target classes.
Because distributed datasets with partially overlapping features are seldom publicly accessible, researcher often simulate distributed conditions by leveraging dataset with overlapping features.

The Indian Liver Patient Dataset (ILPD) from the University of California at Irvine (UCI) Machine Learning Repository \cite{asuncion2007uci, misc_ilpd_225} contains patient data for liver disease. Medical data, especially blood levels, as well as the Age and Gender of the patients, are included. This ILPD data set consists of 579 samples with nine features. The predicted value is encoded as 0 and 1 for liver disease \cite{ramana2011critical}.
The patients and their diagnosis with hepatocellular carcinoma (HCC), a hepatocellular carcinoma, are recorded. Also, data obtained from a control group is contained. Features include, amongst gender, Age, Height, and Weight, three serological biomarkers. The predicted variable, representing hepatocellular carcinoma, is encoded using the values 0 and 1 \cite{best2016galad}. 
The BCD dataset was retrieved from the hospital of the University of Wisconsin \cite{wolberg1990multisurface, misc_breast_cancer_wisconsin__17}. Predictive attributes were gathered from a digital image depicting a fine needle aspirate of a breast mass. Measurements are taken for the features of every cell nucleus visible in the images. The outcome variable pertains to the classification of breast tumors into benign and malignant.

A measure for the imbalance of data sets is the imbalance ratio (IR), which compares the number of minority class samples to the majority class samples \cite{garcia2012effectiveness}. 
The IR is calculated by dividing the number of samples in the minority (Nmin) by the number of samples in the majority class (Nmaj). 

\[ IR = \frac{N_{min}}{N_{maj}} \]

A perfectly balanced data set, that comprises the same amount of samples for both classes, results in an IR of 1. The IR diverges more and more in the direction of zero, the higher the class imbalance gets.
A summary of the data sets’ IR can also be found in Table  \ref{tab:datadescription}.

\subsection{Evaluation}

The evaluation was done in two perspectives, different number of sites contributing to the global model and feature dropped and amount of shared feature ratio. We utilized different datasets (ILPD, HCC, BCD) for benchmark of the federated random forest. When the number of sites is fixed, only the number of features dropped is varied. When the number of features dropped is fixed, the number of sites is varied.
For each run, datasets were split into multiple sites using stratified split, and features are randomly dropped in each sites.
Local RF are built and committed to global federated RF model. Later local RF request aggregated trees from global federated RF and update its local RF. Overview is shown in Figure \ref{fig:overview}.

The significance is evaluated by using the paired t-test and the Wilcoxon signed-rank test as heuristics and verifying the results graphically using the \texttt{DABEST} package \cite{ho2019moving}.
The evaluation takes care of calculating the p-values of the two-sided and one-tailed tests for each metric. The two-sided p-values are extracted from the \texttt{DABEST} package’s analysis code. The one-tailed p-values are calculated using the \texttt{scipy stat} package.

\begin{figure*}
    \centering
    \includegraphics[width=0.432\textwidth]{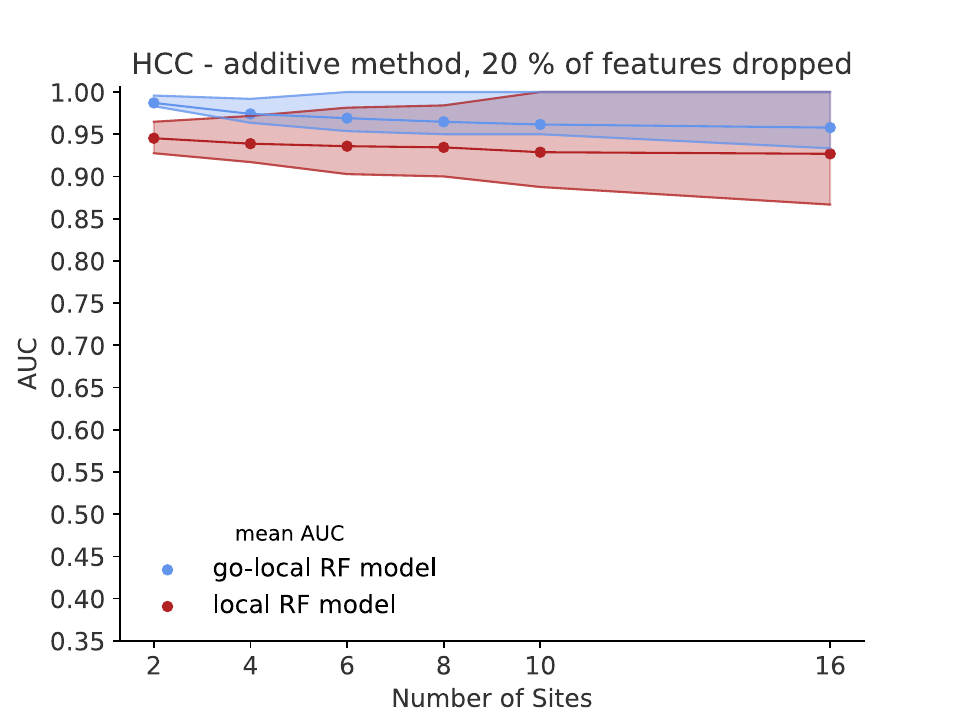}
    \includegraphics[width=0.432\textwidth]{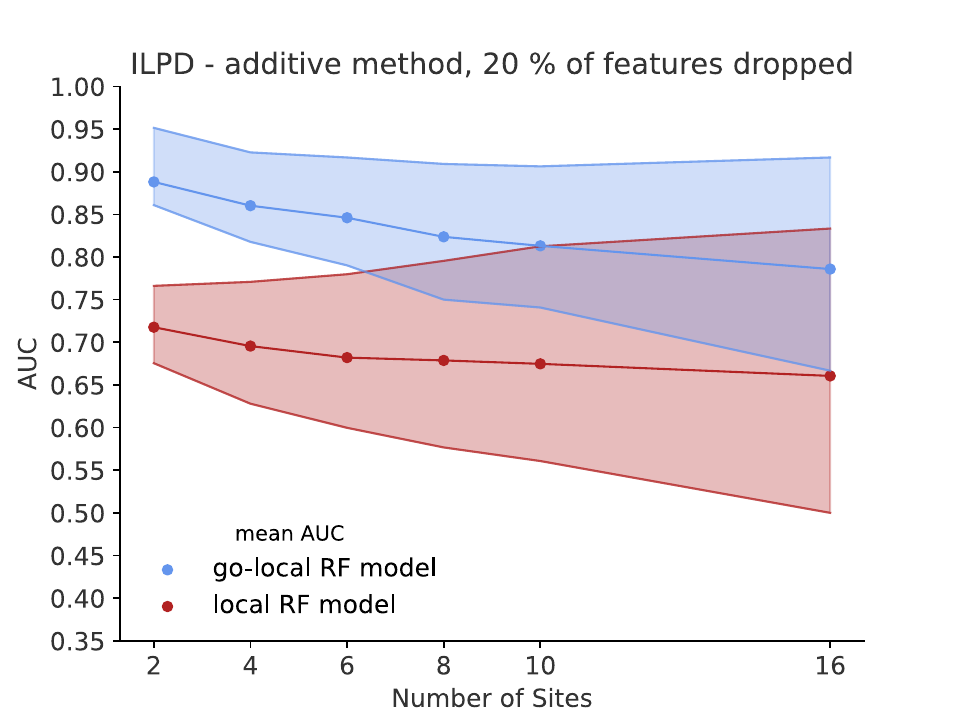}
    \includegraphics[width=0.432\textwidth]{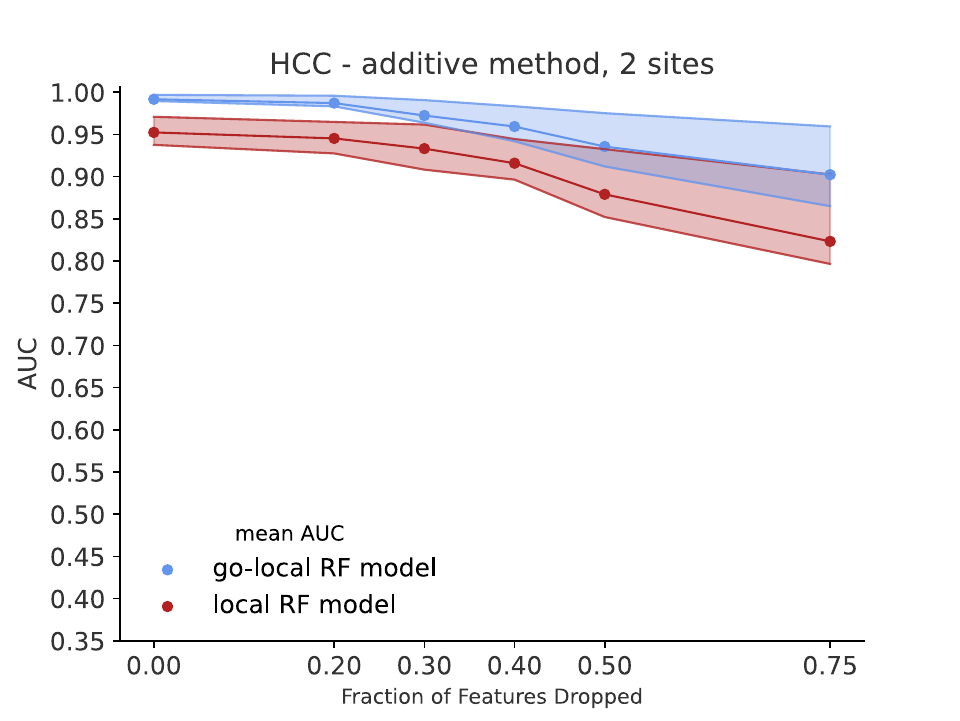}
    \includegraphics[width=0.432\textwidth]{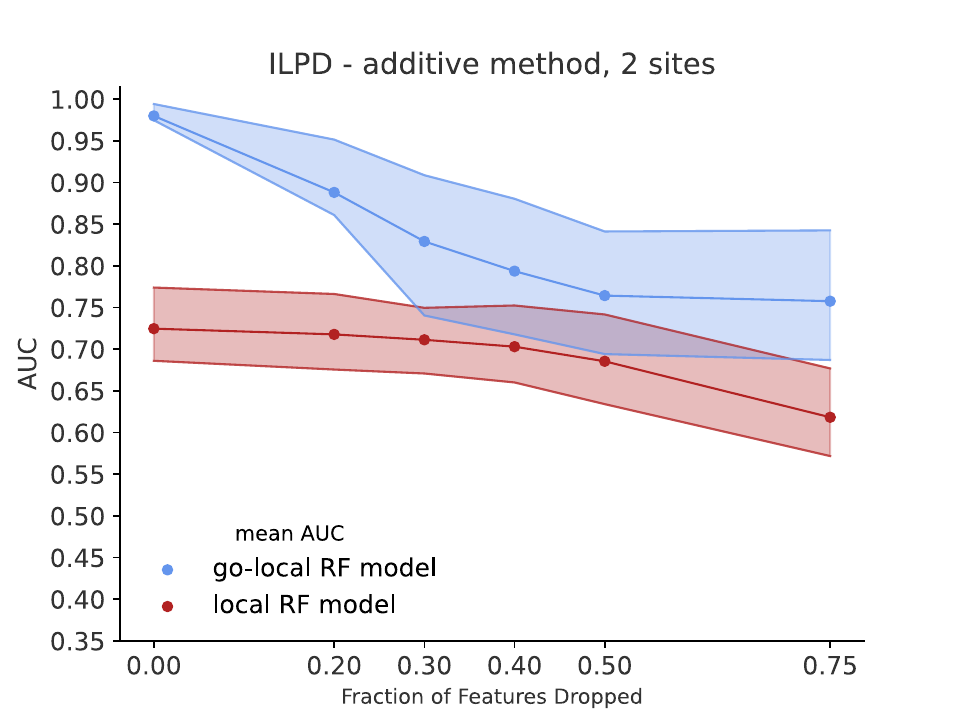}

    \caption{The inner quartile range plots for HCC and ILPD data. Comparison of random forest model improved with federated learning (blue; go-local RF) and local random forest model (red; local RF) is shown. 
    The plots in the first row show AUC scores by varying number of sites when data has 20\% dropped features.
    The plots in the second row show AUC scores by varying number of features dropped when data is split into two sites. 
    The inner quartile range plot surrounds the mean AUC with the respective inner quartile range shaded around.}
    \label{fig:difference_no_of_sites}
\end{figure*}

\subsection{Performance metrics}
To measure a model’s performance in a binary classification task, the inherent characteristics of the data need to be considered. Due to the imbalance of classes in federated learning, performance measures accounting for imbalanced data are suitable. 
The area under the receiver operating characteristics curve (AUC) is a measure barely affected by class imbalance. Nonetheless, the area under the precision and recall curve (PRAUC) varies, which leads to the conclusion that AUC overestimates performance in some cases. Therefore, both measures should be regarded side-by-side. The receiver operating characteristics (ROC) curve and the precision-recall (PR) curve is calculated based on the confusion matrix of the prediction.
For imbalanced data, Matthews correlation coefficient (MCC) is reliable in comparison to accuracy \cite{chicco2020advantages}. The MCC calculates Pearson’s correlation coefficient between actual and predicted classifications and thus ranges from -1 to 1, with a value of -1 depicting the worst classification possible, 0 depicting random guessing, and +1 a perfect prediction achieved by the underlying classifier. The MCC values a correct positive classification in a same way as a correct negative classification.

\section{Results}

In the following sections, the results present the comparison between the local random forest (local RF) and the globally optimized local model (go-local RF). The results presented herein comprise comparisons across three distinct aspects: variation in the number of sites, variation in dropped features, and aggregation method.

\subsection{Variation of the number of sites}

With three datasets, HCC, ILPD, and BCD, globally optimized local models and local models are compared by varying number of sites. Since the original data are split up into multiple sites, more sites consequently comprise fewer samples. 
Figure \ref{fig:difference_no_of_sites} shows the  performance of the analysis on the HCC and ILPD datasets.  
The globally optimized model (go-local RF model) is displayed in blue, whereas the local model is displayed in red.
In general, it can be observed that the globally optimized models perform better than local models. 
Difference between federated model to local model ranged from 0.03$\sim$0.09 in HCC data, 0.075$\sim$0.250 in ILPD data, and 0$\sim$0.01 in BCD data. 

The number of sites in the federation can impact the accuracy of the federated model \cite{hauschild2022federated}. 
With partially overlapping data, similar results were obtained. The effects of increasing data distribution (2 - 16) showed a similar pattern for all datasets.
When the data was divided into more sites, the mean AUC of the go-local RF model decreased significantly compared to that of the local models. 
Furthermore, the increasing variance was observed by increasing number of sites. This could be also explained by decreasing size of data in each sites.
In addition to that, when features are only partially overlapped to each site, the effect of number of sites becomes marginalized.
For instance, in ILPD data, when all features are kept, the go-local RF model is influenced by number of sites. The mean AUC scores of go-local RF changes 97\% (number of sites: 2) to 82\% (number of sites: 16).
As more features were dropped, the mean AUC score of the go-local RF model plateaued across different number of sites. For instance, when 40\% of features were removed, the mean AUC score remained constant at 79\% (with 2 sites) and 78\% (with 16 sites). The same pattern was observed with HCC data.
It is plausible that a substantial reduction in feature overlap could notably diminish the benefits associated with federation.
The PRAUC comparison showed similar trends with AUC results.

\subsection{Variation of the fraction of features dropped}

With three datasets, HCC, ILPD, and BCD, globally optimized local models and local models are compared by varying the number of features dropped.
The mean AUC metrics of the globally optimized model stay above the mean metrics for local models.
This is also supported by the differences displayed in Figure \ref{fig:difference_no_of_sites}.
The discrepancy between globally enhanced models and local models fluctuates with the proportion of dropped features. Nonetheless, no consistent trends were evident across diverse datasets.
On the HCC dataset, the AUC difference between local models and globally optimized models increases as a greater fraction of the features are dropped.
ILPD data showed lowest difference when 50\% features dropped, and BCD data showed less difference as more features dropped.
The PRAUC comparison showed similar trends with AUC results.

Similarly to the examination of the different graphs for the number of sites, it can also be observed that the variance within the graphs increases with the increasing fraction of features that are dropped.
In the HCC dataset, it can be seen that an increase in the number of features removed invariably leads to lower performance.
For the ILPD dataset, a similar trend can be observed%
, but for both 4 sites and 8 sites performance increases for the global-optimized local model when the number of dropped features increases from 50\% to 75\%. Potentially, differently relevant features have been removed here.

\subsection{Aggregation Method}

\begin{figure*}[t]
    \centering
    \begin{subfigure}[b]{0.3\textwidth}
        \centering
        \includegraphics[width=0.95\textwidth]{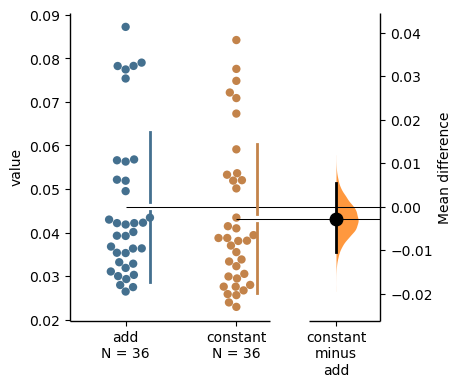}         \caption{HCC}
        \label{fig:HCC_GA_plots}
    \end{subfigure}
    \hfill
    \begin{subfigure}[b]{0.3\textwidth}
        \centering
        \includegraphics[width=\textwidth]{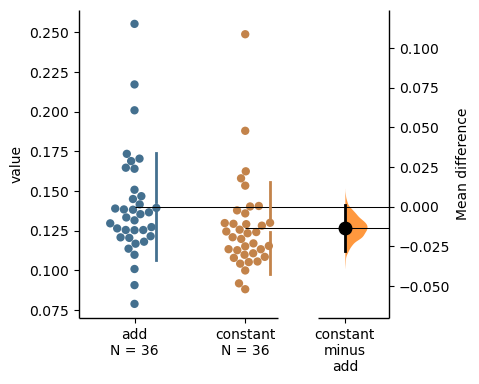}
        \caption{ILPD}
        \label{fig:ILPD_GA_plots}
    \end{subfigure}
    \hfill
    \begin{subfigure}[b]{0.3\textwidth}
        \centering
        \includegraphics[width=\textwidth]{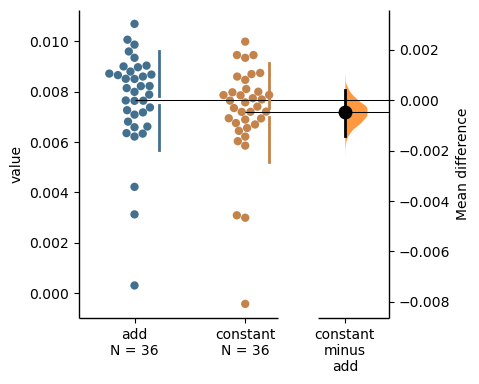}
        \caption{BCD}
        \label{fig:BCD_GA_plots}
    \end{subfigure}

    \caption{Difference of the AUC score between the federation methods.
    The addition and constant method are compared in terms of mean AUCs in all scenarios with different number of sites and features dropped. The Gardner-Altman comparison plots depict the distribution of various split/feature-drop scenarios and the corresponding mean differences among those scenarios.
    }
    \label{fig:GAplots}
\end{figure*}

\begin{table}[b]
    \centering
    \begin{tabular}{c|p{2.2cm}| p{2.2cm}|p{1cm}}
    \toprule
        Data & Mean diff. global-local additive & Mean diff. global-local constant & p-value U-test\\
    \midrule
        HCC & .0459 & .0432  & .168 \\
        ILPD & .1399 & .1268 & .007\\
        BCD & .0077 & .0072 & .062 \\
    \bottomrule
    \end{tabular}
    \caption{Comparison of the mean AUC improvements of the globally optimized model overall parameter combinations for the additive and the constant method.
    The distribution of the difference between local RF and globally optimized local RF is compared.
    }
    \label{tab:difference_AUC_p_value}
\end{table}

Concerning the two aggregation methods - additive and constant - mean AUC and PRAUC of globally optimized models are more often significantly greater than local ones when using the additive method.
Additionally, the difference between the globally optimized models’ metrics seems to be higher for the additive method, in comparison to the constant method. 

The mean differences between the globally optimized model’s and the local model’s AUC for each method with all parameter combinations are shown in Table \ref{tab:difference_AUC_p_value}.
The mean difference averages the performance improvement through global optimization achieved over all parameter combinations. 
In addition, the Mann-Whitney U test (U-test), a non-parametric test that enables significant testing without assuming a specific data distribution, is shown.
Finally, it is examined if the performance improvements of the additive method are significantly greater than the ones of the constant method.
The noticeable contrast between constant and additive is evident solely in the ILPD dataset. 
Although HCC and BCD also show differences between the additive and constant method, the differences were not significant.
This suggests that in situations where there is a considerable variance between locally and globally optimized models, additive may surpass constant aggregation.

\section{Discussion}

This study explores the potential of the newly developed extension of the federated random forests model to a scenario where only a partial overlap exists among features across participating sites. Our major goal is to assess the effectiveness of this algorithm to enhance local predictions within this unique context. 
In contrast to other research that employs transfer learning for knowledge transfer across domains, this study concentrates on the adaptation of a federated random forest model within the context of partially overlapping features. In short, the globally optimized model incorporates additional information about locally available features through the integration of matching trees.
The adapted FRF has been evaluated on three different clinical data sets of different contexts, namely ILPD, HCC and BCD. Finally, the performance is evaluated using the AUC and, because of the imbalance of the data sets, the PRAUC measure.

\textbf{Hypothesis 1: The globally optimized federated models outperform local random forest models when applied to clinical data with partially overlapping features.}
The results confirm our hypothesis that the globally optimized model generally outperforms the respective local model, even with a high number of features dropped for each participating site. 
Thus, random forests are eligible to improve model performance for horizontally partitioned data,
not only for the circumstance where all features are available at all sites but also in a federated setting with  partially overlapping features. 
These improvements can even be achieved without incorporating transfer learning methods.

Moreover, the analysis reveals varied model performance across datasets. Models trained on the BCD dataset consistently perform well with high AUC values close to 1.00 but exhibit lower variance, suggesting possible overfitting.
In contrast, models on the HCC dataset show lower AUC values with larger variance, especially as the number of sites and fraction of dropped features increase.
The ILPD dataset displays significant performance differences, which diminish with more dropped features and sites. 
However, in summary, the globally optimized local models generally outperform local models alone, but the extent of this advantage varies by dataset and target variable. Despite some data overlap, using globally optimized local models can improve performance.

\textbf{Hypothesis 2: With less overlap, there is less improvement in model performance between local and globally optimized models}
In summary, the mean AUCs fluctuated strongly and thus do not support the hypothesis. The result of the analysis of the BCD data only showed a slight decrease in the mean AUC score. The result with ILPD data showed substantial decreases until 50\% features were dropped. However, when 75\% of features were dropped, there was a slight surge of the mean AUC scores.
The HCC result showed an opposite trend with increasing mean AUC score, but lower-scale level.
Therefore, with respect to this hypothesis, it is better to focus on variances of mean AUCs. It is clear that when more features are dropped, the variances of local and globally optimized models are dramatically increased, and the performance improvement from federated learning becomes insignificant.
This is possibly caused by the decreased number of trees in the globally optimized model. Thus, not necessarily all information from the matches is transferred into the optimized model.
This is in agreement with results from Liu \textit{et al.} \cite{liu2020secure}. In their work, the weighted F1 scores are similar for local and federated transfer learning models with a small number of overlapping features. With a growing number of overlapping features, the difference increases.

Although the performance of local models and globally optimized models decreases as more features dropped, in most cases there is still a significant improvement when using the globally optimized model. In the case of the constant method, a slight deterioration of the performance measures was presented. 
When it comes to the aggregation methods, the additive method seems to be more reliable in comparison to the constant method, providing a higher number of significant improvements. Additionally, the difference between globally optimized models and local models increased when using the additive method in comparison to the constant method.

In summary, in this work, the proposed federated random forest is evaluated with respect to the number of sites, the fraction of features dropped, and the aggregation method. 
However, the effect of parameter variations of the random forest model itself has only been examined to a limited extent.
Other important challenges in federated learning, such as privacy, communication, reliability of clients, or non-IID data need to be addressed in future work.
In conclusion, the federated random forest for partially overlapping data can improve predictions of the local model, even with minor feature overlap. 
The healthcare sector encounters persistent challenges in data analysis, primarily stemming from strict data privacy regulations and the presence of varied, non-standardized data, which significantly hinder large-scale analysis efforts.
Therefore, the algorithms showcased here will facilitate the transfer of technology towards clinical practice.





\bibliographystyle{cas-model2-names}

\bibliography{cas-refs}

\end{document}